\documentclass[11pt,a4paper]{article}
\usepackage[hyperref]{emnlp-ijcnlp-2019}
\usepackage{times}
\usepackage{latexsym}
\usepackage{microtype}
\usepackage{todonotes}
\usepackage{url}
\usepackage{graphicx}
\usepackage{amsmath}
\usepackage{amsfonts}
\usepackage{dirtytalk}
\usepackage{bbm}
\usepackage{xcolor}
\usepackage{pifont}
\usepackage{tabularx}
\usepackage{booktabs}
\usepackage{multirow, tabu}

\usepackage[T5,T2A,T1]{fontenc}
\usepackage[utf8]{inputenc}

\definecolor{niceGreen}{rgb}{46,139,87}

\aclfinalcopy 

\title{Semi-supervised Bootstrapping of Dialogue State Trackers \\ for Task Oriented Modelling}

\author{Bo-Hsiang Tseng$^{1}$, Marek Rei$^{2,4}$, Pawe{\l} Budzianowski$^{1,3,5}$ \\   \textbf{Richard E. Turner$^{1}$, Bill Byrne$^{1}$, Anna Korhonen$^{3}$}\\
   ${}^1$Engineering Department,   Cambridge University, UK \\
  ${}^2$Computer Laboratory,  Cambridge University, UK\\
    ${}^3$Language Technology Lab,   Cambridge University, UK\\
  ${}^4$Department of Computing, Imperial College London \\
  ${}^5$PolyAI Limited, London, UK\\
  \texttt{bht26@cam.ac.uk}\\
  }

\date{}

\begin{document}
\maketitle
\begin{abstract}
Dialogue systems benefit greatly from optimizing on detailed annotations, such as transcribed utterances, internal dialogue state representations and dialogue act labels.
However, collecting these annotations is expensive and time-consuming, holding back development in the area of dialogue modelling.
In this paper, we investigate semi-supervised learning methods that are able to reduce the amount of required intermediate labelling. 
We find that by leveraging un-annotated data instead, the amount of turn-level annotations of dialogue state can be significantly reduced when building a neural dialogue system.
Our analysis on the MultiWOZ corpus, covering a range of domains and topics, finds that 
annotations can be reduced by up to 30\% while maintaining equivalent system performance.
We also describe and evaluate the first end-to-end dialogue model created for the MultiWOZ corpus.
\end{abstract}

\section{Introduction}
Task-oriented dialogue models aim at assisting with well-defined and structured problems like booking tickets or providing information to visitors in a new city \cite{raux2005let}. Most current industry-oriented systems rely on modular, domain-focused frameworks \cite{young2013pomdp,sarikaya2016overview}, with separate components for user understanding \cite{henderson2014third}, decision making \cite{gavsic2010gaussian} and system answer generation \cite{wensclstm15}. Recent progress in sequence-to-sequence (seq2seq) modelling has enabled the development of fully neural end-to-end architectures, allowing for different components to be optimized jointly in order to share information \cite{wen2016network,zhao2017learning,budzianowski2019hello}.

Dialogue systems benefit greatly from optimizing on detailed annotations, such as turn-level dialogue state labels \cite{henderson2014third} or dialogue actions \cite{rieser2011reinforcement}, with end-to-end architectures still relying on intermediate labels in order to obtain satisfactory results \cite{lei2018sequicity}.
Collecting these labels is often the bottleneck in dataset creation, as the process is expensive and time-consuming, requiring domain and expert knowledge \cite{asri2017frames}.
Due to this restriction, existing datasets for task-oriented dialogue are several orders of magnitude smaller compared to general open-domain dialogue corpora \cite{lowe2015ubuntu,henderson2019repository}.

Arguably, one of the most challenging parts of dialogue modelling is maintaining an interpretable internal memory over crucial domain concepts \cite{williams:16}. 
Although there is increasing research effort to learn Dialogue State Tracking (DST) jointly with the text generation component \cite{eric2017key,wu2019global}, the most effective models use it as an intermediate signal \cite{wen2016network, lei2018sequicity}. 
The difficulty of state tracking has made this task a driving force behind most of the Dialog System Technology Challenges in recent years \cite{henderson2014third, kim2017fourth}.

\begin{figure*}[!t]
\centering
  \includegraphics[width=0.75\linewidth]{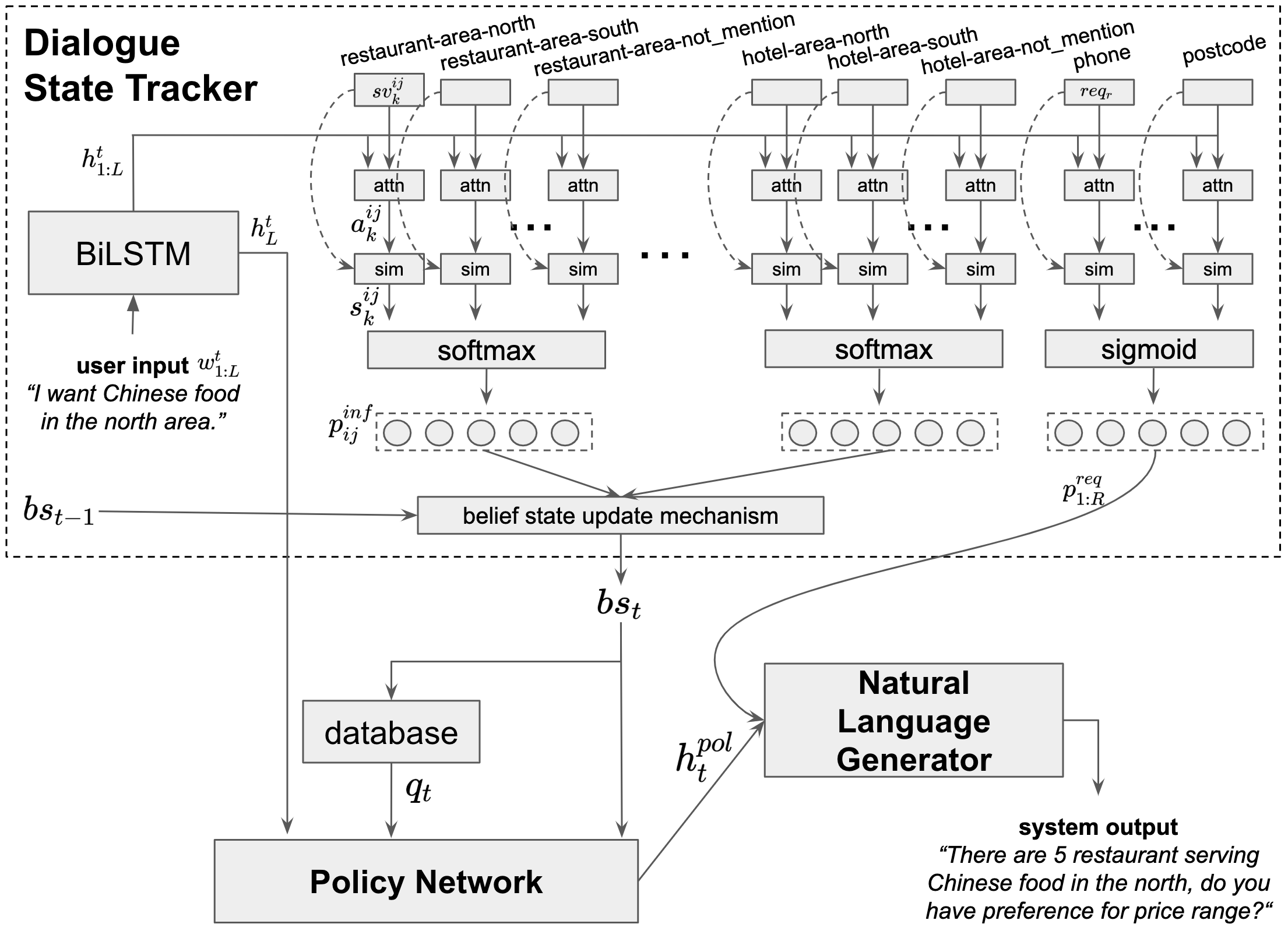}
  \caption{Overview of our end-to-end neural dialogue model. It is composed of three main components: Dialogue State Tracker, Policy Network and Natural Language Generator.}
  \label{fig:e2e}
\end{figure*}

In this paper, we reduce the reliance of task-oriented dialogue systems on data collection by leveraging semi-supervised training \cite{chapelle2009semi}. 
Two approaches are investigated and evaluated for providing an improved training signal to the dialogue state tracking component in an end-to-end dialogue system. 
Automatically predicted DST output on unlabelled utterances is treated as additional annotation if the model confidence is sufficiently high. 
Furthermore, subtle perturbations of existing datapoints are created, optimizing for their predictions to be similar to the original instances.
Our analysis on the MultiWOZ corpus \cite{budzianowski2018multiwoz}, covering a range of domains and topics, finds that these methods can reduce intermediate annotation by up to 30\% while maintaining equivalent system performance.
We also describe and evaluate the first end-to-end dialogue model created for the MultiWOZ corpus.

\section{End-to-end Neural Dialogue Model}
We now present the end-to-end neural dialogue model composed of three main components: dialogue state tracker, policy network and natural language generator. These components are trained jointly as a connected network and will be introduced in detail in the next paragraphs.
The overall architecture can be seen in Figure \ref{fig:e2e}.

\paragraph{Dialogue State Tracker (DST)}
The DST is responsible for both understanding the input user utterance and updating the internal dialogue state for the downstream components.
There are two types of slots which can be detected in the input: informable slots and requestable slots. The former describes the attributes of the entity that the user is looking for, e.g., \texttt{pricerange} of a hotel. The latter captures information that the user desires to know about the entity, e.g., \texttt{postcode} of a hotel.
Each informable slot contains several possible values with two special labels: \texttt{not-mentioned} and \texttt{don't-care}.
A good DST should be able to correctly recognize the mentioned slot-value pairs in the user utterance and to maintain the updated dialogue (belief) state.

Let $i$, $j$ and $k$ denote the index of domain, slot and value.
As depicted at the top of Figure \ref{fig:e2e}, the user utterance {$w_{1}$:$w_{L}$} at turn \textit{t}  is first encoded by the BiLSTM to obtain the hidden states $h^{t}_{1:L}$.
The encoding of the slot-value pair $sv^{ij}_{k}$ is the output of the affine layer that takes the concatenation of the embeddings of domain $i$, slot $j$ and value $k$ as the input.
The context vector $a^{ij}_{k}$ is then computed by the attention mechanism, denoted as \texttt{attn} in Figure \ref{fig:e2e}, following \citet{luong2015effective}:
\begin{equation}
    e_{l} = \text{sim}(h_{l}, sv^{ij}_{k})
\end{equation}
\vspace{-0.5cm}
\begin{equation}
    a^{ij}_{k} = \sum^{L}_{l=1} e_{l} h_{l},
\end{equation}
where $l$ is the word index of the user utterance and $\texttt{sim}$ denotes any function that calculates the similarity of two vectors.
We adopt here the dot product function, following \citet{mrkvsic2017neural,zhong2018global,ramadan2018large}.
The similarity score $s^{ij}_{k}$ between $a^{ij}_{k}$ and $sv^{ij}_{k}$ is then computed to see whether the slot-value pair $sv^{ij}_{k}$ is mentioned in the utterance. The mentioned pair should have higher similarity score to its context vector than those which are not mentioned. The softmax layer is then applied to form the probability distribution $p^{inf}_{ij}$ for each informable slot $s^{inf}_{ij}$, where the predicted value is the value with the highest probability.
The same attention mechanism is used for each requestable slot $req_{r}$ to decide whether the user has asked for the slot in the current turn.
The sigmoid layer is used instead as it is a binary classification problem.
The prediction of requestable slots will be used as input to the natural language generator.

The belief state is the concatenation of the distributions over all informable slot-value pairs that is updated at each turn to keep track of the information provided by the user during the entire conversation.
To form the belief state $bs_{t}$ at turn $t$, for each informable slot $s^{inf}_{ij}$ the update mechanism checks if the predicted value is either \texttt{not-mention} or \texttt{dont-care} for the current turn. If it is, then the probabilistic distribution $p^{inf}_{ij}$ in $bs_{t-1}$ is kept, otherwise it is updated by the new distribution $p^{inf}_{ij}$ at the current turn.


\paragraph{Policy Network}
The policy network is responsible for fusing the signals from the belief state $\mathbf{b}_{t}$, the encoding of the user utterance $\mathbf{h}^{t}_{L}$ and the database query result $\mathbf{q}_{t}$. The database query is constructed from the predicted belief state. The number of all entities that match the predictions of the DST form the database query vector\footnote{Following \cite{wen2016network}, by querying the database using $\mathbf{b}_t$ as query, $\mathbf{q}_{t}$ is the 1-hot representation with each element indicating different number of the matching entities in the database. We use 5 bins to indicate the matching number from 0 to 3 and more than 3.}.
We use a simple feedforward layer as the policy network:

\begin{equation}
   \mathbf{z}_{t} = \text{tanh}(W^{z}*[\mathbf{b}_{t}, \mathbf{h}^{t}_{L}, \mathbf{q}_{t}]).
\end{equation}

where [*] denotes the concatenation of vectors.
\paragraph{Natural Language Generator}
Taking the input $\mathbf{z}_t$ from the policy network and predictions of requestable slots from the tracker, the generator outputs a system response word-by-word until the <EOS> token is generated. To improve the generation of correct slots corresponding to the user input, we adopt the semantically-conditioned LSTM \cite{wensclstm15} that contains a self-updating gate memory to record the produced slots in the generated sentence.

\paragraph{Optimization}
The model is optimized jointly against two sources of information -- DST intermediate labels and system utterances. The DST loss consists of the cross-entropy over the multi-class classification of informable slots while binary cross-entropy is used for requestable slots:
\begin{equation}
    L_{dst}=-\sum_{i}\sum_{j}t^{inf}_{ij}\log p^{inf}_{ij} - \sum_{r}t^{req}_{r}\log p^{req}_{r},
\label{loss:1}
\end{equation}
where $i$, $j$ and $r$ are the index of domain, informable slot and requestable slot respectively; $t^{*}$ is the target distribution. The generation error is a standard cross-entropy between the predicted words and target words:
\begin{equation}
    L_{gen} = -\sum_{l'}t_{l'}\log p_{l'},
\label{loss:2}
\end{equation}
where $l'$ is the word index in the generated sentence.
To jointly train the DST, the policy network, and the generator as a connected network, the final objective function becomes $ L = L_{dst} + L_{gen}$.

\paragraph{Semi-supervised Training}
The DST loss requires each turn to be manually annotated with the correct slot-value pairs.
We experiment with two different semi-supervised training methods that
take advantage of unlabelled examples instead, allowing us to reduce the amount of required annotation.
The first approach is based on the \emph{pseudo-labelling} strategy by \citet{chapelle2009semi}. If the prediction probability of an unlabelled data point for a particular class is larger than a given threshold $\nu$, the example is included in the DST loss with the predicted label. The $\nu$ parameter is optimized on the validation set during development.

The second semi-supervised technique investigated is the \emph{$\Pi$-model} \cite{sajjadi2016regularization} where the input is perturbed with random noise $\epsilon \sim \mathcal{N}(\mathbf{0}, \mathbb{\sigma})$. The perturbations are applied to both labelled and unlabelled data points at the level of embedding of user utterance.  
The model is then required to produce similar predictions ${p}^{inf}_{ij}$ over the belief state compared to the original input, optimized with an additional loss:
\begin{equation}
\label{loss:3}
L3 = \alpha \frac{1}{N} \sum_N \sum_{ij} (t^{inf}_{ij} - {p}^{inf}_{ij})^2,
\end{equation}
\noindent where $N$ is the batch size and $\alpha$ is a hyperparameter controlling the weight of the loss. 


\begin{figure}[tb]
  \includegraphics[width=\linewidth]{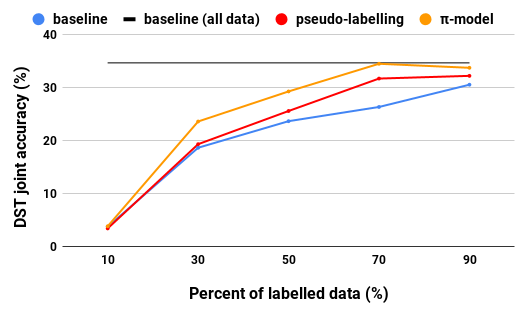}
  \caption{The DST joint accuracy for the three considered models as the amount of labelled data varies. The horizontal line denotes the baseline model trained on 100\% labelled data.}
  \label{fig:dst_res}
    \vspace{-1em}
\end{figure}

\section{Experiments} 
We investigate the effects of semi-supervised training on optimizing the end-to-end neural dialogue system.
In particular, we evaluate how much annotation at the intermediate-level could be reduced while preserving comparable results of the overall dialogue task completion metrics. 

\paragraph{Dataset}
The three analyzed models are evaluated on the MultiWOZ dataset consisting of 10,438 task-oriented dialogues \cite{budzianowski2018multiwoz}. The conversations in MultiWOZ are natural as they were gathered based on human-to-human interactions following the Wizard-of-Oz paradigm. The corpus includes multi-domain conversations spanning across $7$ domains including Restaurant, Hotel, Attraction, Train and Taxi. The size of the dataset allows us to control the amount of available fully labelled datapoints.

\paragraph{Metrics} 
There are two metrics of importance when evaluating task-oriented dialogue systems. The first is the DST joint goal accuracy, defined as an average joint accuracy over all slots per turn \cite{williams:16}. The second is the \emph{Success} metric that informs how many times systems have presented the entity satisfying the user's goal and provided them with all the additional requested information \cite{wen2016network}. The models are optimized using the validation set and the results are averaged over 10 different seeds.

\paragraph{Varying data amount}
\begin{figure}[tb]
  \includegraphics[width=\linewidth]{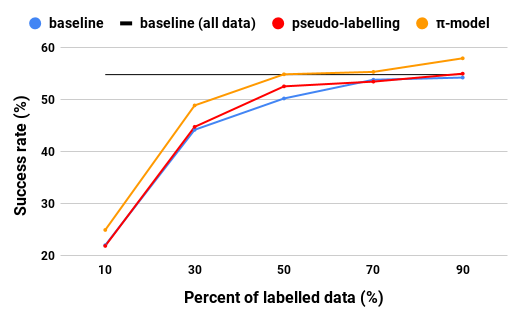}
  \caption{Success rate for different methods as the amounts of labelled data varies. Horizontal line denotes the baseline model trained on 100\% labelled data.}
  \label{fig:success_res}
  \vspace{-1em}
\end{figure}

We examine the performance of the baseline model compared to the two semi-supervised models as the amount of labelled data varies. The result of the DST joint accuracy is presented in Figure \ref{fig:dst_res}. The \emph{pseudo-labelling} model performs better than the baseline when more than 50\% of the dataset is labelled. At the scarce data levels (10\% and 30\%) , the \emph{pseudo-labelling} model is not producing pseudo training points that help improve DST predictions. In contrast, the \emph{$\Pi$-model} takes advantages of the additional regularization loss and effectively leverages unlabelled data to enhance the performance over the baseline.
The improvements are consistently more than 5\% when training with 30 to 90\% of labelled data and even reach the performance of the fully trained baseline model with only 70\% labelled data.

Figure \ref{fig:success_res} shows the \textit{Success} metric results. The \emph{pseudo-labelling} method is not able to improve performance over the baseline regardless of the amount of labelled data. However, the \emph{$\Pi$-model} is capable of improving the success rate consistently and manages to reach the performance of the fully trained model with only 50\% of the intermediate DST signal. Note that a better DST joint accuracy does not necessarily translate to a better success rate as the final metric is also influenced by the quality of the generator.

\begin{table}[h!]
\centering

\resizebox{\columnwidth}{!}{%
\begin{tabu}{l||ccccc}
\tabucline [1pt]{1-6}
\textbf{No. of examples} & \textbf{0}   & \textbf{1-5}   & \textbf{6-10}  & \textbf{10-15} & \textbf{16-20} \\
\tabucline [1pt]{1-7}
\textbf{Baseline}       & 6.17         & 15.93          & 25.71          & 35.07          & 28.88         \\
\textbf{Pseudo-labelling}              & 6.5 & 16.27 & 26.96 & 33.46 & 28.55 \\
$\mathbf{\Pi}$-\textbf{model}              & \textbf{6.6} & \textbf{21.93} & \textbf{31.29} & \textbf{36.22} & \textbf{30.72} \\
\tabucline [1pt]{1-6}

\end{tabu}%
}
\caption{The accuracy (\%) of different classification of slot-value pairs in terms of their number of training examples.} 
\label{tab:sv_res}
\end{table}

\paragraph{DST analysis} DST joint accuracy considers all slot-value pairs in an utterance and cannot give us further insight regarding the source of the improvements. We are particularly interested in whether the semi-supervised models can leverage unlabelled data to improve the prediction of rarely seen slot-value pairs. In this analysis, we classify all slot-value pairs in the test set in terms of their number of training examples in 50\% of the labelled data. Table \ref{tab:sv_res} presents the results, showing that the \emph{$\Pi$-model} improves accuracy by 5\%  when the slot-value pair is rarely (1-10 times) seen during training. The improvement on few-shot slots contributes to the improvement of joint accuracy.

\paragraph{Domain analysis}
\begin{figure}[tb]
  \includegraphics[width=\linewidth]{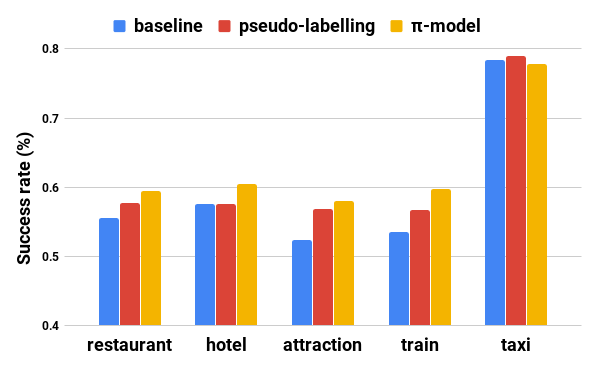}
  \caption{Success rates in each domain in the case of 50\% labelled data.}
  \label{fig:domain_res}
\vspace{-1em}
\end{figure}
We also investigate if the improvements in the success rate are consistent among all domains. Figure \ref{fig:domain_res} shows the success rate on individual domains 
in the case of 50\% of data is labelled. Both semi-supervised models improve performance over the baseline in all domains except for the taxi domain. We hypothesize this comes from the fact that the taxi domain is a relatively easy domain with only $4$ possible slots. 

\section{Conclusions and Future Work}
In this paper, we have analyzed how much semi-
supervised techniques could help to reduce the
need for intermediate-level annotations in training neural task-oriented dialogue models. The results suggest that we do not need to annotate all intermediate signals and are able to leverage unannotated examples for training these components instead.
In the future, we plan to experiment with other intermediate signals like dialogue acts. Further improvements
could potentially be obtained from employing more advanced regularization losses \cite{oliver2018realistic}.

\section*{Acknowledgments}
Bo-Hsiang Tseng is supported by Cambridge Trust and the Ministry of Education, Taiwan. Marek Rei's research is supported by Cambridge English via the ALTA Institute. Pawe{\l} Budzianowski is funded by an EPSRC grant (EP/M018946/1) and by Toshiba Research Europe Ltd, Cambridge  Research  Laboratory (RG85875).

\bibliography{acl2019}
\bibliographystyle{acl_natbib}



\end{document}